\documentclass[10pt,twocolumn,letterpaper]{article}

\usepackage[pagenumbers]{cvpr} %

\usepackage{graphicx}
\usepackage{amsmath}
\usepackage{amssymb}
\usepackage{booktabs}

\usepackage[pagebackref,breaklinks,colorlinks]{hyperref}

\usepackage{xcolor}
\usepackage{xspace}

\usepackage{subcaption}
\usepackage{multirow}
\usepackage{tabularx}
\usepackage{listings}
\usepackage{soul}

\long\def\ignorethis#1{}

\newcommand{\MASK}{\texttt{[MASK]}\xspace}
\newcommand{\SMASK}{\texttt{[*]}\xspace}
\newcommand{\W}{$\left<w\right>$\xspace}
\newcommand{\SW}{$\left<s\right>$\xspace}

\definecolor{tomato}{RGB}{255,99,71}
\definecolor{gold}{RGB}{255,215,0}
\definecolor{mediumaquamarine}{RGB}{102,205,170}
\definecolor{blueviolet}{RGB}{138,43,226}
\definecolor{cornflowerblue}{RGB}{100,149,237}
\definecolor{lightpink}{RGB}{255,182,193}
\definecolor{orchid}{RGB}{218,112,214}

\newbox\jsavebox
\newcommand{\jsubfig}[2]{%
	\sbox\jsavebox{#1}%
	\parbox[t]{\wd\jsavebox}{\centering\usebox\jsavebox\\#2}%
	}

\newcommand{\SEP}{\texttt{[SEP]}\xspace}

\usepackage[capitalize]{cleveref}
\crefname{section}{Sec.}{Secs.}
\Crefname{section}{Section}{Sections}
\Crefname{table}{Table}{Tables}
\crefname{table}{Tab.}{Tabs.}

\def\cvprPaperID{5058} %
\def\confName{CVPR}
\def\confYear{2023}

\begin{document}

\title{Is BERT Blind? Exploring the Effect of Vision-and-Language Pretraining on Visual Language Understanding}

\author{Morris Alper$^*$, Michael Fiman$^*$, Hadar Averbuch-Elor\\
Tel Aviv University}

\twocolumn[{%
\renewcommand\twocolumn[1][]{#1}%
\maketitle
\thispagestyle{empty}
\begin{center}
\vspace{-4pt}
  \centering
        \jsubfig{\includegraphics[height=2.85cm]{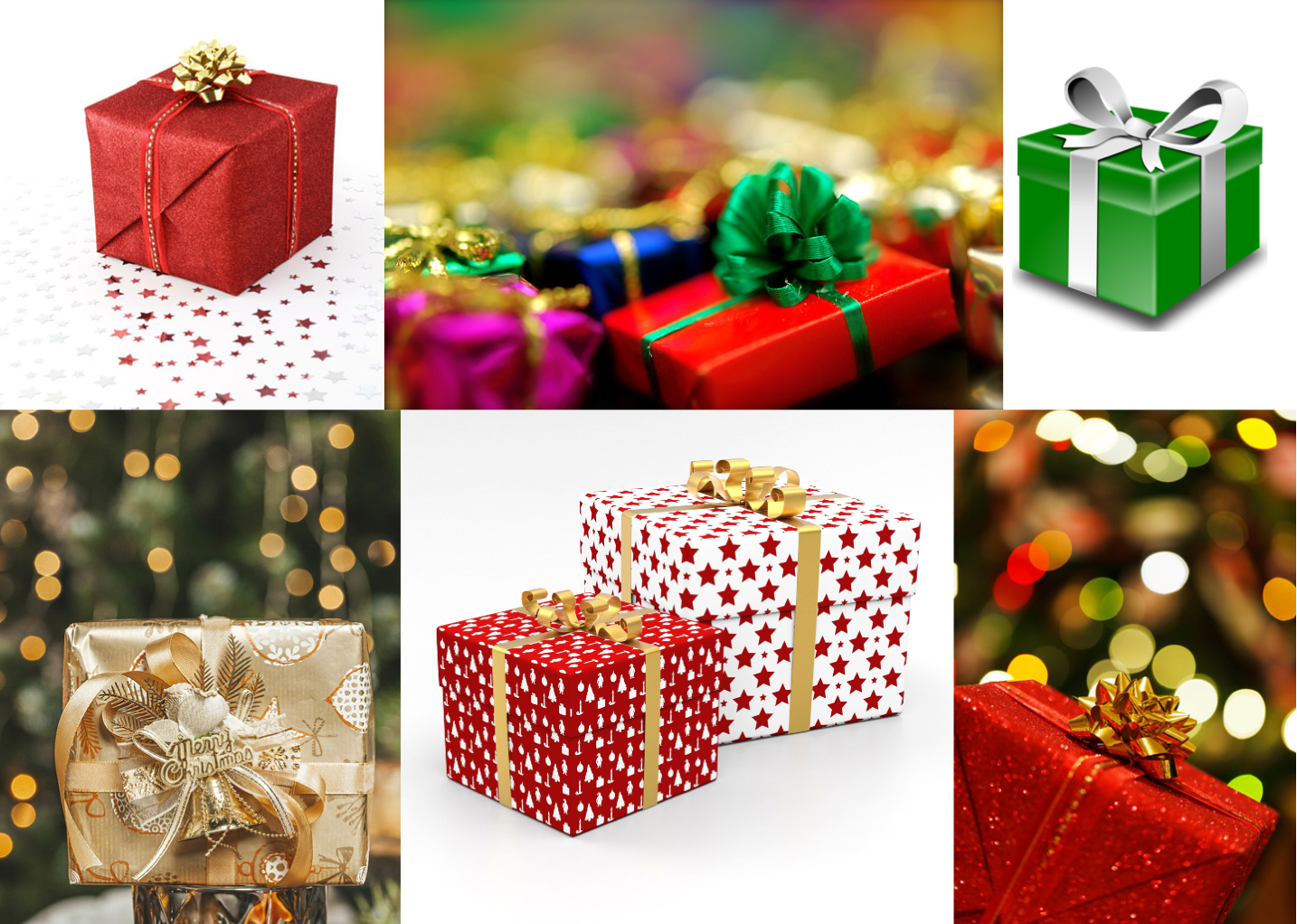}}{Alex gave Riley a present}
    \hfill
    \jsubfig{\includegraphics[height=2.85cm]{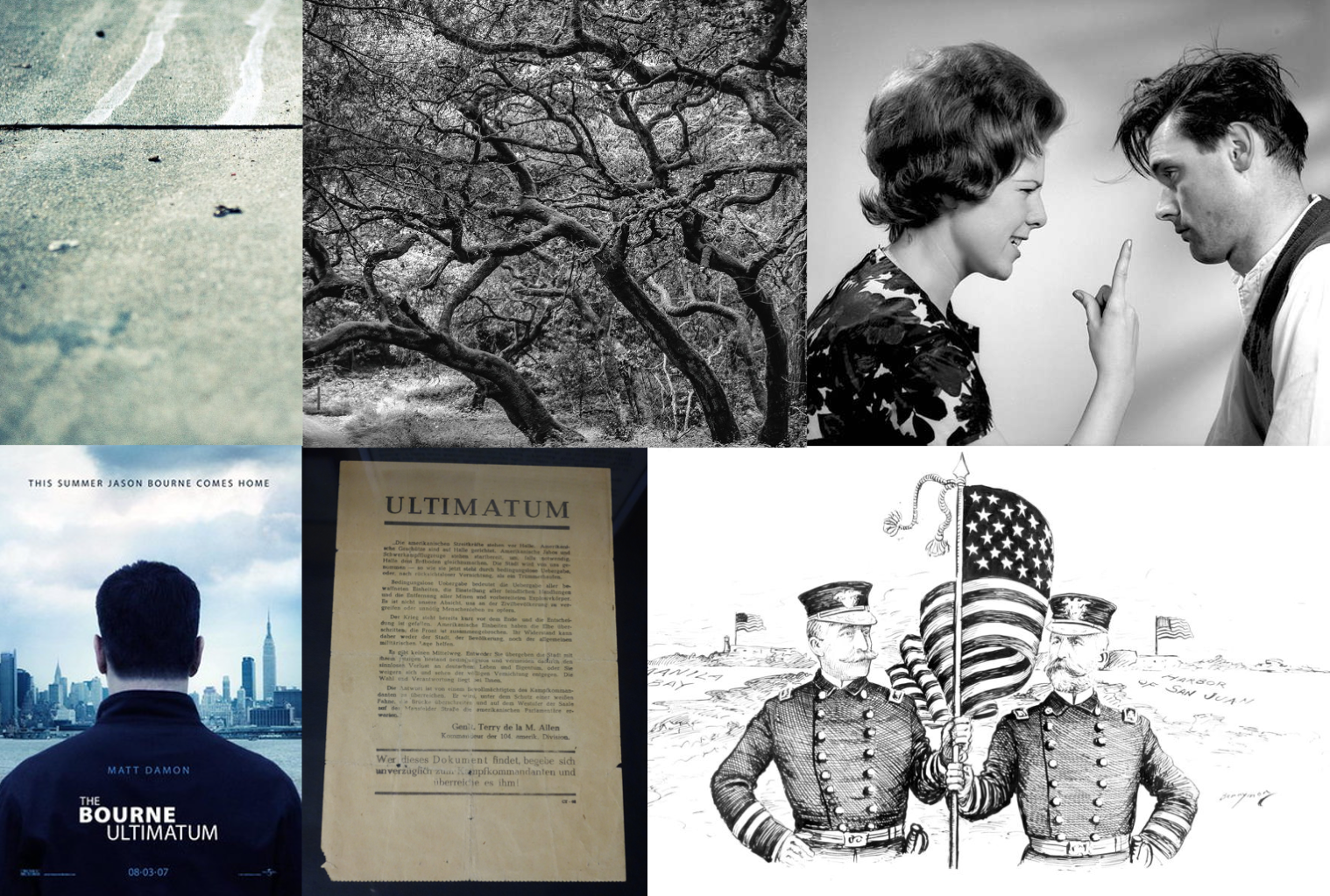}}{Alex gave Riley an ultimatum}
    \hfill \hfill \hfill \jsubfig{\includegraphics[height=2.85cm]{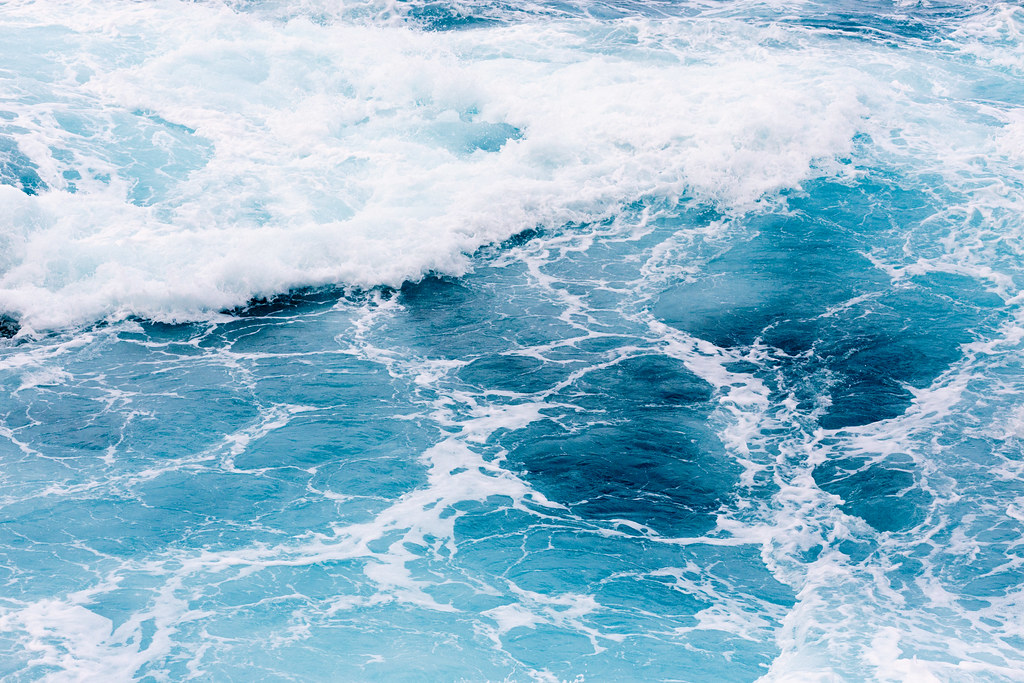}}{the ocean is \SMASK-colored}
    \hfill
    \jsubfig{\includegraphics[height=2.85cm]{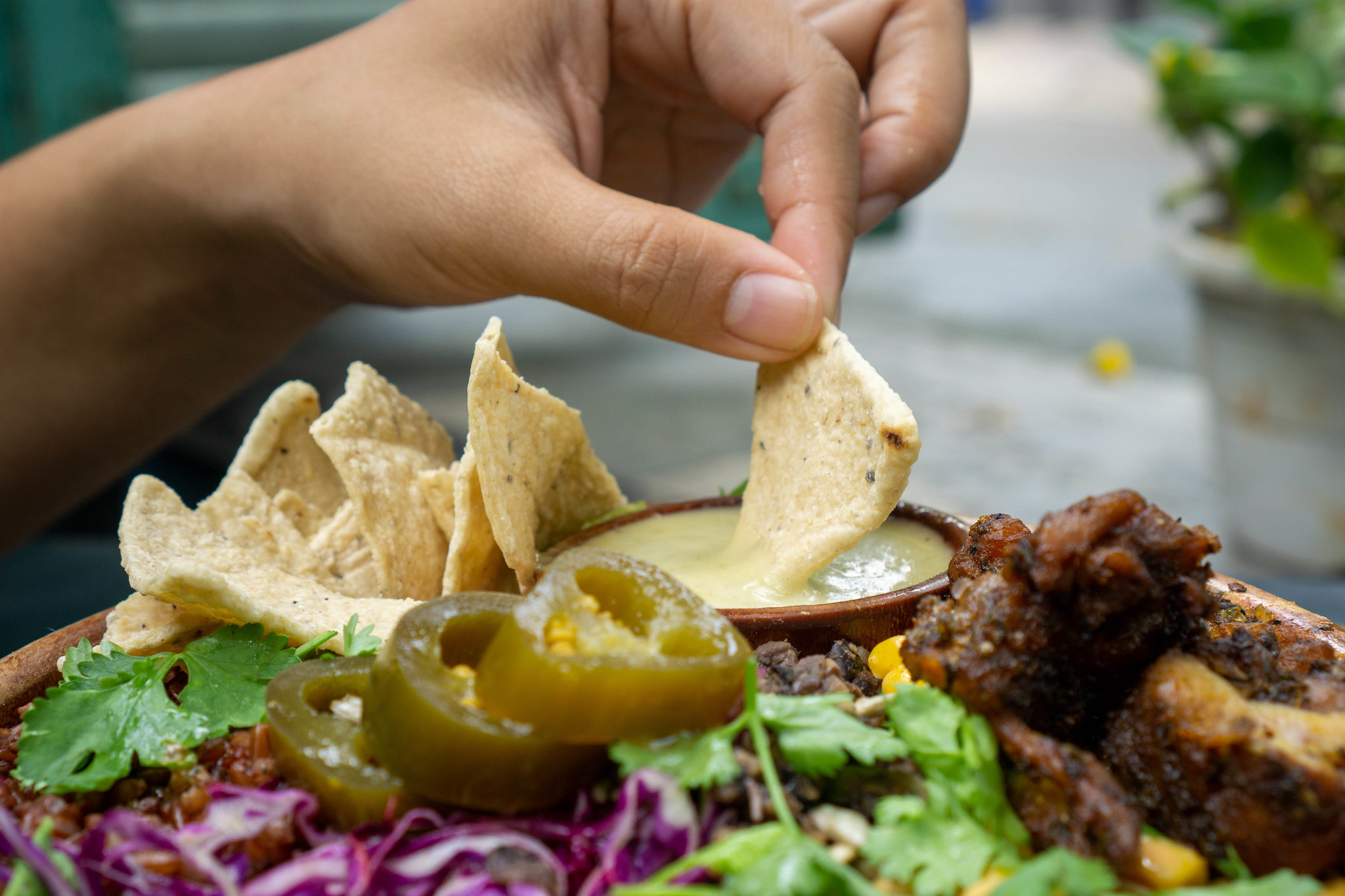}}{a corn chip is \SMASK-shaped}
    \begin{tabularx}{\textwidth}{cc}
    \hspace{11pt}\textbf{How concrete are the words \emph{present} and \emph{ultimatum}?} & \hspace{24pt} \textbf{What word should be inserted in the blank?}
    \end{tabularx}
    
    \captionof{figure} {In this paper, we propose a suite of \emph{visual language understanding} tasks for probing the visual reasoning capabilities of text encoder models. While we consider text-only tasks (i.e., processing only the textual descriptions above without associated imagery), we argue that they require visual imagination to complete and can thus benefit from vision-and-language pretraining. For instance, consider the words \emph{present} and \emph{ultimatum}. A simple online query (that considers only freely-available images) roughly yields a coherent set of images for the more concrete word (namely \emph{present}), while the latter cannot be uniquely depicted. Likewise, selecting the most natural color or shape descriptors in cloze contexts as shown in the two examples on the right requires implicit knowledge of the appearance of the referent under consideration (\emph{ocean} and \emph{corn chip} respectively).
    }
  \label{fig:teaser}
\end{center}
}]

\def\thefootnote{*}\footnotetext{These authors contributed equally to this work}\def\thefootnote{\arabic{footnote}}

\begin{abstract}
   Most humans use visual imagination to understand and reason about language, but models such as BERT reason about language using knowledge acquired during text-only pretraining. In this work, we investigate whether vision-and-language pretraining can improve performance on text-only tasks that involve implicit visual reasoning, focusing primarily on zero-shot probing methods. We propose a suite of visual language understanding (VLU) tasks for probing the visual reasoning abilities of text encoder models, as well as various non-visual natural language understanding (NLU) tasks for comparison. We also contribute a novel zero-shot knowledge probing method, Stroop probing, for applying models such as CLIP to text-only tasks without needing a prediction head such as the masked language modelling head of models like BERT. We show that SOTA multimodally trained text encoders outperform unimodally trained text encoders on the VLU tasks while being underperformed by them on the NLU tasks, lending new context to previously mixed results regarding the NLU capabilities of multimodal models. We conclude that exposure to images during pretraining affords inherent visual reasoning knowledge that is reflected in language-only tasks that require implicit visual reasoning. Our findings bear importance in the broader context of multimodal learning, providing principled guidelines for the choice of text encoders used in such contexts\footnote{Our code will be made available at \url{https://isbertblind.github.io/}}.

\end{abstract}

\section{Introduction}
Humans are multimodal learners. We communicate with each other about things that we have experienced and knowledge we have gained using our senses---most commonly including sight as well as hearing, touch, smell, and taste. Our communication channel is limited to a single modality---spoken language, signed language, or text---but a reader or listener is expected to use his or her imagination to visualize and reason about the content being described. In general, language is used to describe scenes, events, and images; the words used to describe these are used to conjure up a visual impression in the listener. Therefore, it is natural to consider the types of visual reasoning used in understanding language, and to ask how well we can currently model them with computational methods.

Consider, for instance, the questions in Figure \ref{fig:teaser}. Concreteness is typically correlated with how well a concept can be visually imagined. For example, a concrete word such as \emph{present} often has a unique visual representation. In addition, common associations such as \emph{ocean}$\rightarrow$\emph{blue} (color) and \emph{corn chip}$\rightarrow$\emph{triangle} (shape) reflect properties of an imagined visual representation of the item in question. These properties may be difficult to infer from text alone without prior knowledge gained from visual input; for instance, a number of studies have investigated the partial ability of blind English speakers to predict color associations and how it differs from the intuition of sighted speakers\footnote{This phenomenon is illustrated in \href{https://www.youtube.com/watch?v=59YN8_lg6-U}{this interview} with Tommy Edison, a congenitally blind man, in which he describes his understanding and frequent confusion regarding color associations.}~\cite{van2021blind, saysani2021seeing, saysani2018colour, shepard1992representation, marmor1978age}. 

There has been a wealth of recent research vision-and-language (V\&L) tasks involving both text and image data, and the use of vision-language pretraining (VLP) to create models that are able to reason jointly about both of these modalities together~\cite{chen2020uniter,kim2021vilt,li2021align,chen2022vlp}. Notable in this regard is CLIP~\cite{radford2021learning}, consisting of paired text and image encoders jointly trained on a contrastive objective, that learns to align text and image embeddings in a shared semantic space. On the other hand, text encoder models such as BERT~\cite{devlin2018bert} learn to reason about text in a unimodal vacuum, with knowledge derived from pretraining tasks that only involve textual data.

Prior work has investigated the performance of multimodally trained text encoders on various natural language understanding (NLU) tasks with mixed results, sometimes finding that they are outperformed by unimodal models~\cite{iki2021effect} and at other times suggesting improved performance~\cite{wang2021simvlm}. However, these works fine-tune the models under consideration on NLU tasks before evaluation, making it difficult to disentangle the effects of multimodal pretraining and fine-tuning configuration on the observed performance. Additionally, these works do not address the distinction between NLU tasks requiring implicit visual reasoning and ones that are purely non-visual. We refer to natural language inference involving implicit visual reasoning as \emph{visual language understanding} (VLU) and propose a suite of VLU tasks that may be used to evaluate visual reasoning capabilities of pretrained text encoders, focusing primarily on zero-shot methods.

We compare multimodally trained text encoders such as that of CLIP to BERT and other unimodally trained text encoders, evaluating their performance on our suite of VLU tasks. We evaluate these models in without modifying their internal weights in order to probe their knowledge obtained during pretraining. A key design aspect of these tests is the probing method used to evaluate knowledge. Previous work has probed the knowledge of BERT and similar models using a masked language modelling (MLM) paradigm~\cite{petroni2019language, rogers2020primer}, but this cannot be directly applied to CLIP since it was not pretrained with MLM. We therefore propose a new zero-shot probing method that we term \emph{Stroop probing}. This is based on the psychological Stroop effect~\cite{macleod1991half} (described in Section \ref{sec:probing}), which suggests that salient items should have a stronger interference effect on the representation of their context.

Strikingly, we find that the multimodally trained text encoders under consideration outperform unimodally trained text encoders on VLU tasks, both when comparing to much larger encoders as well as ones of comparable size. We also compare these models on baseline NLU tasks that do not involve visual reasoning and find that models such as CLIP underperform on these tasks, demonstrating that they do not have a global advantage on NLU tasks. We conclude that exposure to images during pretraining improves performance on text-only tasks that require visual reasoning. Furthermore, our findings isolate the effect of the text component of multimodal models for tasks such as text to image generation, providing principled guidelines for understanding the knowledge that such models inject into downstream vision tasks.

\section{Related Work}
Building models to create text embeddings for a large range of
language tasks has been broadly explored over the past several years. In our work we compare two types of transformer-based models which encode these types of embeddings: those trained on text-only data (unimodally trained), and those exposed to both text and image data during training (multimodally trained).

Since the introduction of the self-attention-based transformer architecture by Vaswani \emph{et al.}~\cite{vaswani2017attention}, transformers have become the predominant architecture for tasks involving textual data. Devlin et al.\cite{devlin2018bert} introduce BERT, a transformer model encoding contextual information for each token in an input sequence in a bidirectional manner. They suggest a self-supervised pretraining method to compute contextual feature representations of textual data, via masked language modelling and next sentence prediction objectives. This pretrained model can then be applied to other downstream tasks by end-to-end fine-tuning. Subsequently, various other text encoder transformers have been proposed, such as RoBERTa~\cite{liu2019roberta}, DistilBERT~\cite{sanh2019distilbert}, and ERNIE~\cite{sun2019ernie, sun2020ernie}. While these models differ on some architectural details and on precise pretraining objectives, they all share the basic transformer architecture and the use of denoising pretraining objectives.  In particular, they are all trained on unimodal data, meaning that they are only exposed to text during training.

In contrast to unimodally trained text encoders, V\&L models have been exposed to both text and image data during training. These models are typically used for tasks that require joint reasoning on text and images, such as visual question answering, grounding referring expressions, and vision-language retrieval \cite{chen2022vlp,gan2022vision}. Fusion encoder models such as LXMERT~\cite{tan2019lxmert}, UNITER~\cite{chen2020uniter}, ViLT~\cite{kim2021vilt}, and ALBEF~\cite{li2021align} output a fused representation of text and image data, while dual encoder models like CLIP~\cite{radford2021learning} and ALIGN~\cite{jia2021scaling} consist of dual text and image encoder models that are jointly trained to produce embeddings in a shared semantic space. FLAVA, a vision-and-language model introduced recently by Singh et al. \cite{singh2022flava}, also includes dual text and image encoders, and is trained on both multimodal objectives involving the alignment of images and text, as well as unimodal objectives. In this work we focus on the text encoder component of dual encoder models, since it may be used after V\&L pretraining for text-only tasks.

Various works have explored the use of multimodal learning to benefit text understanding.
Most related to our work is the very recent study of Zhang et al.~\cite{zhang2022visual} which investigates the use of unimodal and multimodal models for understanding visual commonsense in text. 
Their analysis follows a line of related work investigating the contribution of multimodal learning to visual commonsense knowledge in text, since such knowledge is typically not written explicitly in text but is abundantly present in visual information~\cite{vedantam2015learning,lin2015don,kottur2016visual}. Unlike Zhang et al.~\cite{zhang2022visual} who only evaluate CLIP with an added set of task-specific learned weights, we are able to probe CLIP and other similar models in the strictly zero-shot setting via our novel Stroop probing method. This allows for directly evaluating properties learned by the models, independent of differences that result, for instance, from specific training configurations. In addition, %
we also study performance on both visual and non-visual NLU tasks in order to provide a controlled benchmark.

Other works have investigated the use of multimodal learning for NLU in various contexts. Bruni et al.~\cite{bruni2014multimodal} propose an architecture for integrating text and image-based distributional information to improve performance on tasks where meaning is grounded in perception. Kiela and Bottou~\cite{kiela2014learning} show that integrating features extracted from images using CNN with skip-gram representation vectors improves performance on semantic similarity datasets. Lazaridou et al.~\cite{lazaridou2015combining} train visual representations extracted using CNNs together with skip-gram embeddings to integrate visual and textual information performing well in both semantic and vision tasks. Kiela et al.~\cite{kiela2017learning} train a sentence embedding model using grounded information extracted from image features by attempting to predict the image features. These embeddings improve performance on various NLP tasks in comparison to text only embeddings. They show that using this method for a dataset consisting mainly of abstract words is likely to less benefit from grounding information.  Shi et al.~\cite{shi2019visually} show that a syntactic parser can benefit from seeing images during training; however, it was later shown that the model mostly relies on noun concreteness (which we also elaborate on in our work) rather than more complex syntactic reasoning~\cite{kojima2020learned}.
The use of images for PCFG induction is also investigated by Jin \& Schuler~\cite{jin2020grounded}.

Along with the rise in visibility of jointly trained V\&L transformer models, a number of works have explored the use of these models for text-only tasks, with mixed results. Using terms coined by Sileo et al.~\cite{sileo2021visual}, these can be broadly split into \emph{associative grounding} and \emph{transfer grounding} approaches. Associative grounding uses retrieval methods to associate particular images with related texts; Kiros et al.~\cite{kiros2018illustrative} and Tan \& Bansal~\cite{tan2020vokenization} show that associative grounding methods may improve performance on various text-only NLU benchmark tasks. Transfer grounding applies V\&L models directly to text-only input, disregarding the vision component of the model during inference. Wang et al.~\cite{wang2021simvlm} apply this to weakly-supervised V\&L models to outperform BERT on various text-only tasks from the GLUE benchmark. On the other hand, Iki \& Aizawa~\cite{iki2021effect} find that V\&L-pretrained text encoders have similar or inferior results on NLU tasks including tasks from GLUE. Likewise, Cao et al.~\cite{cao2020behind} find that although visually-aligned text encoders perform well on semantic linguistic probing tasks, BERT still outperforms them.

As discussed above, some prior works suggest that multimodal pretraining aids text understanding while other works show that it can lead to degradation. Our work provides new context for these seemingly contradictory results, allowing them to be reassessed in the new context of visual vs. non-visual natural language understanding.

\section{Experimental Setup}

\subsection{Models Used}

For evaluating unimodally trained text encoders, we use BERT~\cite{devlin2018bert}, RoBERTa~\cite{liu2019roberta}, DistilBERT and DistilRoBERTa~\cite{sanh2019distilbert}, which are all trained with text-only MLM objectives. We also include results for Sentence-BERT (SBERT)~\cite{reimers2019sentence}, since its output embeddings are trained to have meaningful cosine similarity scores and thus bear more similarity to other models evaluated with Stroop proving. Results on multimodally trained text encoders are reported for CLIP~\cite{radford2021learning} and FLAVA~\cite{singh2022flava}; for these models we use only the text encoder with pretrained weights and discard the other subcomponents. Our tests include checkpoints from both OpenAI and the OpenCLIP open-source implementation of CLIP~\cite{cherti2022reproducible, ilharco_gabriel_2021_5143773}. Details of the checkpoints used for each model are listed in the supplementary material.

The text encoders of the multimodally trained models range in size from 63M (CLIP) to 109M (FLAVA) parameters. We compare to both comparably small unimodally trained text encoders such as DistilBERT (66M parameters) as well as much larger text encoders such as BERT-large (340M). See the supplementary material for an exhaustive list of sizes of the models under consideration.

We use each model with frozen pretrained weights. Our subsequent tests probe the contents of the feature vectors extracted by these models. For MLM probing, we also use the model's MLM head for prediction. In cases where MLM can be used we have found it to outperform Stroop probing; in such cases we report results for MLM probing here and for Stroop probing in the supplementary material.

\subsection{Probing Methods} \label{sec:probing}

In order to probe the inherent knowledge of our models, we use the knowledge probing methods described below. The probing methods that follow are strictly zero-shot; in the supplementary material we analyze the use of linear classifiers trained on our models' frozen embeddings (``linear probing'').

\medskip \noindent \textbf{Masked language modelling (MLM)}. BERT and our other unimodally trained models were all pretrained with MLM objectives and hence can be used for zero-shot prediction of tokens in a masked context. Given a text including a \MASK{} token and a set of $k$ possible completions $C = \{c_1, c_2, \cdots, c_k\}$, a MLM assigns probabilities $p_1, \cdots, p_k$ to each corresponding token. We use $\arg \max_i p_i$ as the model's prediction. Previous works have found that BERT and other MLM can be probed for innate knowledge with this method~\cite{petroni2019language,rogers2020primer}.

\medskip \noindent \textbf{Stroop probing (SP)}. We propose another zero-shot probing method to extract knowledge from models based on the pooled embeddings that they extract. Consider a masked text $t_m$ and possible completions $c \in C$, and let $t_c$ be the text with $c$ inserted in the mask location. Given a text encoder $M$, we calculate pooled embeddings $v_m = M(t_m)$ and $v_c = M(t_c)$ %
and unit-normalize them to $\hat{v}_m = v_m / \|v_m\|$ and $\hat{v}_c = v_c / \|v_c\|$. Stroop probing considers the cosine similarity scores $s_c := \hat{v}_m \cdot \hat{v}_c$. These can be used either directly for regression (as in the concreteness task below), or for categorical prediction by selecting $c^* = \arg \max_c s_c$.

The intuition behind Stroop probing is that items which are more surprising, incongruous, or salient in the given context may have a stronger interference effect on the encoding of the surrounding text. This is analogous to the \emph{Stroop effect} in human psychology. When presented with congruent and incongruent stimuli such as color words printed in the same or differing colors (e.g. ``red'' printed in blue), readers take significantly longer on average to read the incongruent stimuli, a phenomenon known as the \emph{Stroop effect}~\cite{macleod1991half}\footnote{For example, try saying these colors out loud (not the printed words): {\color{red}Green},  {\color{green}Red}, {\color{purple}Blue}, {\color{green}Purple}, {\color{blue}Red}, {\color{red}Purple}. }. We use Stroop probing for multiple tasks, including predicting color associations, as described below.

\subsection{Prompts Used}

For each task, we test the probing methods above on a wide variety of prompts in order to show the robustness of the described phenomena. In our results below we report the maximum metric value for each model over all of the prompts, since this represents a rough bound on our ability to extract intrinsic knowledge from the models under consideration. A full list of prompts used for each task and an analysis of model performance across prompts are provided in the supplementary material.

In some cases our prompt contains an empty slot, which we indicate below as $\SMASK$. Some models under consideration have a dedicated mask token, but for those such as CLIP that do not, we insert a fixed token in this slot, detailed further in the supplementary material.

\section{VLU Tasks}
\label{sec:vlu}

We present three VLU tasks to probe the ability of our models to understand language with implied visual context: concreteness prediction (Section \ref{sec:concreteness}), color association prediction (Section \ref{sec:color}) and shape association prediction (Section \ref{sec:shape}). Note that each of these tasks is performed on text alone, but requires visual reasoning to complete.

\subsection{Concreteness Prediction}
\label{sec:concreteness}

\medskip \noindent \textbf{Task description.} Words and phrases can be roughly classified as either \emph{concrete} or \emph{abstract}. A concrete concept is something that can be pointed to or directly sensed, while abstract concepts refer to things that cannot be easily visualized~\cite{schwanenflugel2013abstract}. This can be conceptualized on a scale, ranging from the most abstract to the most concrete. Psychological research suggests that concrete words are easier for humans to understand and remember than abstract words~\cite{schwanenflugel2013abstract}. Similarly, it has been shown that concreteness correlates with the learnability of visual concepts for machine learning models~\cite{hessel2018quantifying}, and that MLM pretraining of V\&L models may be improved by preferentially masking concrete words~\cite{bitton2021data}.

Because concreteness is a property of text that is tightly coupled with the visual domain, we consider concreteness prediction to be a VLU task, requiring some knowledge of the visual content of language to complete. We note that this task has been addressed in various previous works~\cite{hill2014multi,hill2014learning,hessel2018quantifying,rabinovich2018learning,charbonnier2019predicting}. In contrast to these approaches, our unsupervised concreteness estimation procedure evaluates the concreteness of a word or phrase in a given textual context, rather than being limited to a fixed set of lexical items or discrete categories in a dataset.

\medskip \noindent \textbf{Experimental details.} We probe our models for the concreteness of words in context by using a cloze task paradigm with Stroop probing. For example, using the prompt $t_m = $ \emph{``I see the \SMASK''} and testing word \W, we insert the word into the prompt to obtain $t_w = $ \emph{``I see the \W''}, and use cosine similarity score $s_w$ between embeddings of $t_m$ and $t_w$ as the regression output. All prompts used are listed in the supplementary material. 

We test our approach on the dataset introduced by Brysbaert et al. \cite{brysbaert2014concreteness}. This dataset contains 39,954 English unigrams and bigrams along with human-labelled concreteness scores on a scale from 1 (abstract) to 5 (concrete), averaged over annotators. We only use the unigram nouns from this list, totaling 14,562 items. Note that unlike prior concreteness prediction techniques that train supervised models on this dataset~\cite{charbonnier2019predicting}, we perform zero-shot prediction on this task with no supervised training, using the dataset for testing only.

Also note that we do not report results for DistilBERT or DistilRoBERTa since the checkpoints used do not contain a trained pooling layer, which is required for Stroop probing.

\medskip \noindent \textbf{Evaluation metrics.} We report absolute values of Pearson, Spearman, and Kendall correlations between the predicted concreteness and ground truth scores ($\left|\rho\right|$, $\left|r_s\right|$, and $\left|\tau\right|$ respectively).
\subsection{Color Association Prediction}
\label{sec:color}

\medskip \noindent \textbf{Task description.} Some concepts are highly associated with particular colors---for example, the word \emph{banana} is highly associated with the color yellow, while a word like \emph{child} does not have a strong color association. These color associations have been widely studied in experimental psychology and neuroscience~\cite{bramao2011role,bannert2013decoding}. We propose a task of \emph{color association prediction} -- given a noun (or noun phrase) \W, identify the color with which \W is normally associated.

\medskip \noindent \textbf{Experimental details.} To probe our models for color associations, we use the MLM and SP methods described above. In particular, we conceive of this task as categorical prediction over a set of basic color words $C$. For example, using the prompt \emph{``A picture of a \SMASK \W''} where \W is the item being tested, our probing methods search for the most suitable color to place in the \SMASK slot. All prompts tested are listed in the supplementary material. For MLM probing, we predict the color $c \in C$ with the highest predicted probability in the \SMASK{} slot of the prompt. For SP, we predict $c^* = \arg \max s_c$ using similarity scores as defined above.

To test this method on our chosen text encoders, we use two datasets. The Color Terms Dataset (CTD)~\cite{bruni2012distributional} provides a list of 52 concrete words and their color. The Natural-Color Dataset (NCD) of fruit images~\cite{anwar2020image} is a colorization task containing images of 20 types of fruits and vegetables paired with colors. We use the provided list of fruits and colors as a fixed set of words with strong color associations, discarding the image data. For the latter, we filter objects with the color label \emph{purple} as this label contains multiple WordPiece tokens and thus is not directly comparable with MLM probing for models such as BERT. This results in 15 unique fruits and vegetables. For each model, we calculate color predictions using the probing methods described above out of the set: $\{$\emph{red, orange, yellow, green, blue, black, white, grey, brown}$\}$.

\medskip \noindent \textbf{Evaluation metrics.} We report categorical accuracy of predictions on the CTD and NCD datasets ($acc_{CTD}$ and $acc_{NCD}$) relative to the ground truth color labels.
\subsection{Shape Association Prediction}
\label{sec:shape}

\medskip \noindent \textbf{Task description.} Another salient visual feature of language is the association between concrete nouns or noun phrases and particular shapes. For example, the nouns \emph{wheel} and \emph{compass} have a circular association, while \emph{pyramid} and \emph{corn chip} have a triangular association. Shape associations have been studied in the psychological literature in contexts such as child language acquisition~\cite{yee2011function} and semantic memory representation~\cite{verdine2016shape}. Building on this line of research, we propose the task of \emph{shape association prediction} -- given a noun (or noun phrase) \W, identify the basic shape that is most associated with \W. Because the space of possible shapes is complex and difficult to categorize unambiguously, we restrict \W under consideration to nouns associated with a few basic shapes, as described below.

\medskip \noindent \textbf{Experimental details.} We construct the \textbf{ShapeIt} benchmark for shape associations\footnote{available in our code repository}. This contains 109 items total, each consisting of a noun or noun phrase along with the basic shape most associated with it from the set $\{rectangle, circle, triangle\}$. The benchmark was constructed by performing a user study requiring users to choose a shape associated with a given word, and selecting for only those words which were consistently classified by the users. Data collection methods used in constructing this benchmark are detailed in the supplementary material, along with further analysis of its contents. Probing methods used for this task are equivalent to the color association prediction task. Prompts used for probing include \emph{``A \SMASK shaped \W''} where \W is the shape associated word; the full list of prompts used is detailed in the supplementary material. We use both shape nouns (e.g. \emph{circle}) and associated adjectives (e.g. \emph{circular}) and report the highest accuracy achieved between these two settings.

\medskip \noindent \textbf{Evaluation metric.} We report categorical accuracy of predictions ($acc$) relative to the ground truth shape labels.

\section{Non-visual NLU Tasks}
\label{sec:nlu}

We also present three non-visual NLU tasks to serve as a baseline comparison for our models:

\subsection{Factual Knowledge Probing}
\label{sec:factual}

\medskip \noindent \textbf{Task description.} It has been observed that language models have an emergent \emph{knowledge base} property, in which they may be probed for factual knowledge about the world~\cite{petroni2019language, rogers2020primer}. Various works on probing BERT and other language models for commonsense world knowledge have found that they show an impressive ability to memorize knowledge, although they may be deficient in applying this knowledge to reasoning about the physical world~\cite{forbes2019neural}. In this task, we probe our models for fine-grained factual knowledge via a cloze test, where an empty slot must be filled in with a word. We test on factual knowledge about geographical locations since this requires factual knowledge that does not explicitly rely on visual reasoning.

\medskip \noindent \textbf{Experimental details.} For this task, we use the Comparative Question Completion dataset introduced by \cite{zagoury2021s}. This consists of questions in which one of a pair of coordinated elements is masked; the target is the masked phrase. Specifically, we use the \emph{cities} dataset which masks the names of geopolitical entities such as cities and countries. Example sentences from the dataset include: \emph{which country has more part time jobs new zealand or \SMASK{}?} (the correct answer being \emph{australia}) and \emph{which is older saudi arabia or \SMASK{}? } (the correct answer being \emph{persia}).
The original dataset has 1,187 questions with 447 unique locations as answers. In order to fit the general method of masking tasks, we filter masked phrases with more than one token (e.g. \emph{the west coast}) similar to the protocol presented in the original paper. As this results in an extremely limited set of candidates for MLM models such as RoBERTa that use Byte Pair Encoding tokenization, we restrict the MLM models under comparison to BERT and DistilBERT. The filtered dataset contains 825 questions with 216 unique locations.

We treat this task as a categorical classification task, choosing only from the set of unique locations given in the dataset per sample, and evaluating how often the correct target is chosen. We use MLM probing and Stroop probing for categorical prediction as described above. Similarly to our other tasks, the intuition is that more surprising completions should have a larger interference effect on the text's encoding, if the relevant information is encoded in the embedding.

\medskip \noindent \textbf{Evaluation metrics.} We report recall at one and five ($R@1$, $R@5$), measuring how often the ground truth answer is found among the model's top one or five predicted candidates.
\subsection{Language Proficiency Probing}
\label{sec:proficiency}

\medskip \noindent \textbf{Task description.} In order to evaluate our models' intrinsic knowledge of general language usage, we consider the task of reconstructing English text in order to produce natural-sounding language. Multiple-choice cloze tests are commonly used in language assessment tasks for students to evaluate their proficiency~\cite{stubbs1974cloze,alderson1979cloze,trace2020clozing}. Similarly, a model with a good grasp of English language usage should be able to fill in missing words in cloze contexts to produce fluent English. This requires grammatical and semantic knowledge, but in general, it is not directly related to visual reasoning when applied to arbitrary masked contexts. As noted by Trace~\cite{trace2020clozing}, cloze tasks may evaluate global reading comprehension or local contextual information in the cloze context; we focus on the latter case and refer to this task applied to our models  as \emph{language proficiency probing}.

\medskip \noindent \textbf{Experimental details.} To evaluate language proficiency, we use the Children’s Book Test (CBT) cloze dataset provided by Meta research~\cite{hill2015goldilocks}, consisting of book passages with accompanying masked sentences and possible mask completions. We discard the book passages and only consider the sentences and completions, to focus on the task of reconstructing well-formed text. Completions are grouped by part of speech (POS); we use the noun (N), verb (V), and preposition (P) groups and discard the named entity groups since the latter require long-distance context to predict while N, V, and P can often be inferred from local sentential context. In total, each of the N, V, and P groups contains 2,500 sentences with 10 possible completions each. We filter out long sentences since our multimodally trained models have shorter expected input lengths. After filtering we are left with 1,588 noun, 1,747 verb, and 2,382 preposition completion sentences. In addition, we only use sentences that have a one-word token answer for all tokenizers. For example, one sentence from the V group is \emph{I \SMASK not a fellow; I am a young lady!} and the set of possible completions is \{\emph{am, born, find, picking, pricked, said, sat, seems, streamed, thinking}\}. We use MLM and Stroop probing to evaluate our models on this data.

\medskip \noindent \textbf{Evaluation metrics.} We report categorical accuracy per POS group ($acc_V$, $acc_N$, and $acc_P$), measuring how often the ground truth answer is selected in each of these groups.
\subsection{Sentiment Analysis}
\label{sec:sentiment}

\medskip \noindent \textbf{Task description.} \emph{Sentiment analysis} refers to the task of predicting speaker emotion or affect, a well-studied problem in natural language processing~\cite{jurafsky2023speechchapter4,li2017reflections, poria2020beneath}. We focus on sentiment analysis in text as a subset of text classification. Since text describing the same visual scene may have a positive or negative sentiment (\emph{This cake is delicious} vs. \emph{This cake tastes bad}), we consider this task to be a non-visual NLU task.

\medskip \noindent \textbf{Experimental details.} For this task, we use the IMDB movie review dataset consisting of 50K movie reviews with binary sentiment labels (positive/negative)~\cite{maas2011learning}. In order to provide a fairer comparison between models rather than biasing towards models trained on longer texts, we use only a single random sentence from each review in the IMDB dataset. In addition, we filter long sentences which are too long for multimodal encoders leaving 42,567 examples. Using only a single sentence makes this task more challenging since the randomly chosen sentence is not guaranteed to contain sufficient context for understanding the review's sentiment, but we find that significantly better than random performance is achievable, as seen in the results section. We also differ from the more common learned sentiment analysis paradigm by using strictly zero-shot prediction via engineered prompts. For example, one prompt used is \emph{``sentiment expressed for the movie is \SMASK. \SW''}, where \SW indicates the sentence chosen from the initial review, and \SMASK may be filled with one of $\{$\emph{good, bad}$\}$. We apply MLM and Stroop probing for binary prediction, and report categorical accuracy achieved for each model.

\medskip \noindent \textbf{Evaluation metric.} We report categorical accuracy of predictions ($acc$) relative to the ground truth sentiment labels.
\section{Results and Discussion}

\begin{table*}[t]
  \centering
  \setlength{\tabcolsep}{3.5pt}
  \def\arraystretch{0.95}
  \begin{tabularx}{0.92\textwidth}{lcccccccccccccc}
     \toprule
    &
    \multicolumn{3}{c}{\textbf{Concreteness}} &
    \multicolumn{2}{c}{\textbf{Color}} &
    \textbf{Shape} &&& 
    \multicolumn{2}{c}{\textbf{Knowledge}} &
    \multicolumn{3}{c}{\textbf{Proficiency}} &
    \textbf{Sent.} 
    \\
    \cmidrule(lr){2-4}
    \cmidrule(lr){5-6}
    \cmidrule(lr){7-7}
    \cmidrule(lr){10-11}
    \cmidrule(lr){12-14}
    \cmidrule(lr){15-15}
    Metric & $\left|\rho\right|$ & $\left|r_s\right|$ & $\left|\tau\right|$ &
    $acc_{\texttt{CTD}}$ & $acc_{\texttt{NCD}}$ & acc &&&
    R@1 & R@5 & $acc_{\texttt{V}}$ & $acc_{\texttt{N}}$ & $acc_{\texttt{P}}$ & acc
    \\
    \midrule
    \textbf{Unimodal} \\
    BERT-base & 0.414 & 0.416 & 0.283 & 0.353 & 0.400 & 0.559 &&& 0.198 & 0.522 & 0.898 & 0.753 & 0.893 & 0.618 \\
    BERT-large & 0.348 & 0.355 & 0.239 & 0.490 & 0.467 & 0.587 &&& \textbf{0.231} & \textbf{0.541} & \textbf{0.914} & \textbf{0.779} & 0.905 & 0.625 \\
    DistilBERT & -- & -- & -- & 0.333 & 0.400 & 0.587 &&& 0.148 & 0.479 & 0.864 & 0.709 & 0.814 & 0.637 \\
    RoBERTa-base & 0.433 & 0.404 & 0.275 & 0.431 & 0.333 & 0.431 &&& -- & -- & 0.877 & 0.718 & 0.881 & 0.666 \\
    RoBERTa-large & 0.345 & 0.374 & 0.253 & 0.471 & 0.400 & 0.431 &&& -- & -- & 0.898 & 0.765 & \textbf{0.916} & \textbf{0.703} \\
    DistilRoBERTa & -- & -- & -- & 0.411 & 0.333 & 0.431 &&& -- & -- & 0.804 & 0.664 & 0.756 & 0.635 \\
    ERNIE & 0.461 & 0.496 & 0.338 & 0.196 & 0.333 & 0.449 &&& 0.001 & 0.006 & 0.064 & 0.051 & 0.086 & 0.582 \\
    ERNIE-large &  0.358 & 0.353 & 0.233 & 0.216 & 0.267 & 0.458 &&& 0.006 & 0.022 & 0.209 & 0.241 & 0.280 & 0.674 \\
    SBERT & 0.338 & 0.337 & 0.228 & 0.198 & 0.067 & 0.513 &&& 0.013 & 0.141 & 0.237 & 0.158 & 0.126 & 0.554 \\
    \midrule
    \textbf{V\&L} \\
    CLIP & 0.603 & 0.624 & 0.437 & 0.843 & \textbf{0.800} & 0.798 &&& 0.009 & 0.118 & 0.134 & 0.133 & 0.126 & 0.560 \\
    OpenCLIP & \textbf{0.634} & 0.643 & 0.432 & \textbf{0.941} & \textbf{0.800} & \textbf{0.853} &&& 0.009 & 0.121 & 0.211 & 0.135 & 0.123 & 0.560 \\
    FLAVA & 0.608 & \textbf{0.665} & \textbf{0.449} & 0.882 & \textbf{0.800} & 0.798 &&& 0.020 & 0.138 & 0.116 & 0.118 & 0.139 & 0.519 \\
    \bottomrule
    
  \end{tabularx}
  \vspace{-5pt}
  \caption{\textbf{Results on VLU (left) and non-visual NLU (right) tasks:} concreteness prediction, color and shape association prediction, factual knowledge probing, language proficiency probing, and sentiment analysis (Sent.) respectively. For tasks other than concreteness prediction, MLM probing is used for models supporting it (BERT, DistilBERT, RoBERTa, DistilRoBERTa); SP is used elsewhere. The definition of each metric is defined in the relevant task definition in Sections \ref{sec:vlu}-\ref{sec:nlu}. DistilBERT and DistilRoBERTa do not have concreteness results due to the pooling layer issue mentioned in Section \ref{sec:concreteness}, and RoBERTa and DistilRoBERTa do not have results for factual knowledge probing due to the tokenization issue mentioned in Section \ref{sec:factual}. As these results show, V\&L models yield superior performance on visual tasks, while underperforming unimodally trained models on non-visual NLU tasks.} 
\label{tab:all_results}
\end{table*}

Results for the tasks described above are provided in Table \ref{tab:all_results}. For tasks with multiple prompts the listed metrics are the maximum over prompts, providing a rough upper bound on each model's ability to perform the task in question. Further analysis of performance by prompt, as well as SP results for models shown here with MLM, are provided in the supplementary materials.

As seen in these results, multimodally trained models consistently outperform unimodally trained models on VLU tasks, including both comparably sized and much larger text encoders, while generally underperforming them on non-visual NLU tasks. This is further illustrated by qualitative analysis of the results in various tasks.

\begin{figure}[t]
        \centering
        \begin{subfigure}[t]{.245\linewidth}
            \centering
            \begin{tabular}{c}
\begin{lstlisting}
snowman
liar
lettuce 
mailman
couch
\end{lstlisting}
             \end{tabular}
             \caption*{BERT}
        \end{subfigure}%
        \begin{subfigure}[t]{.245\linewidth}
            \centering
            \begin{tabular}{c}
\begin{lstlisting}
sink
bench
chalk
splinter
pinecone
\end{lstlisting}
\end{tabular}
\caption*{CLIP}
\end{subfigure} 
\begin{subfigure}[t]{.245\linewidth}
            \centering
            \begin{tabular}{c}
\begin{lstlisting}
seed
jelly
cash
lightning
pudding
\end{lstlisting}
\end{tabular}
\caption*{BERT}
\end{subfigure}%
\begin{subfigure}[t]{.245\linewidth}
\centering
\begin{tabular}{c}
\begin{lstlisting}
friend 
story
name
thanks
fun
\end{lstlisting}
\end{tabular}
\caption*{CLIP}
\end{subfigure}
    \textbf{Most \emph{concrete}} \hspace{48pt} \textbf{Most \emph{abstract} }
\caption{Basic nouns selected as most (and least) concrete using BERT-base and CLIP, according the method described in Section \ref{sec:concreteness}. As these illustrate, concreteness can be reasonably predicted from CLIP text embeddings, whereas this knowledge is not readily accessible for the unimodally trained text encoders.}
\label{fig:concreteness_examples}
\end{figure}

\begin{figure}
    \centering
      \jsubfig{\includegraphics[height=1.44cm]{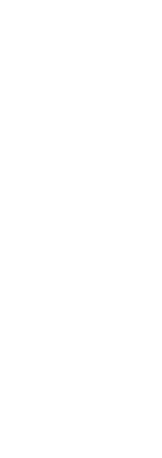}}{\vspace{-22pt}
    \begin{flushleft} BERT     CLIP
    \end{flushleft}} 
  \hspace{10pt}
  \jsubfig{\includegraphics[height=1.44cm]{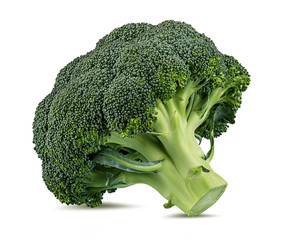}}{green \\         green}
  \jsubfig{\includegraphics[height=1.44cm]{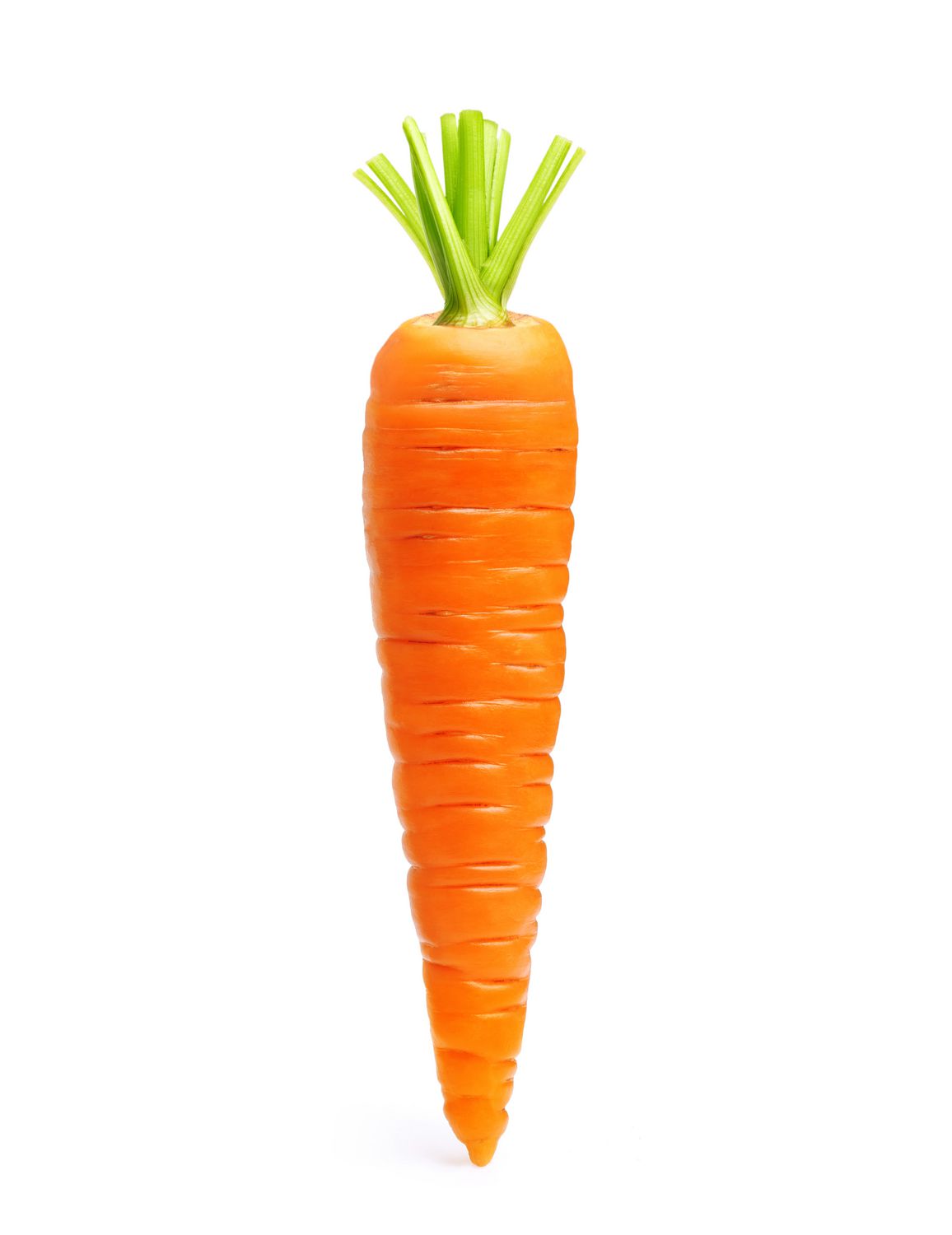}}{green  \\        orange}
  \jsubfig{\includegraphics[height=1.44cm]{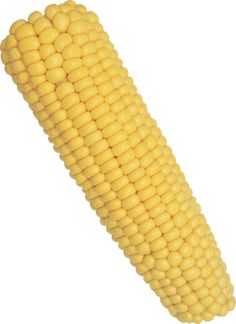}}{red  \\        yellow}
  \jsubfig{\includegraphics[height=1.44cm]{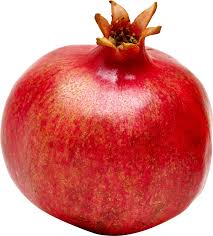}}{red  \\        red}
  \jsubfig{\includegraphics[height=1.44cm]{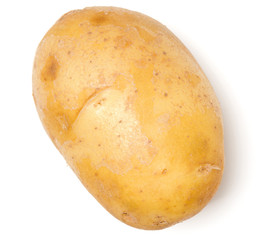}}{white  \\        brown}
    \caption{\textbf{Examples of color prediction results on NCD dataset}. Depicted above are examples of results for predicting colors from the NCD dataset using MLM for BERT-base and Stroop probing for CLIP. We emphasize that the model only receives as input the name of the fruit or vegetable without the given image.}
    \label{fig:colors}
\end{figure}
\begin{figure}
    \centering
      \jsubfig{\includegraphics[height=1.5cm]{figures/color/white.png}}{\vspace{-22pt}
    \begin{flushleft} Item     BERT     CLIP
    \end{flushleft}} 
  \hspace{10pt}
  \jsubfig{\includegraphics[height=1.5cm]{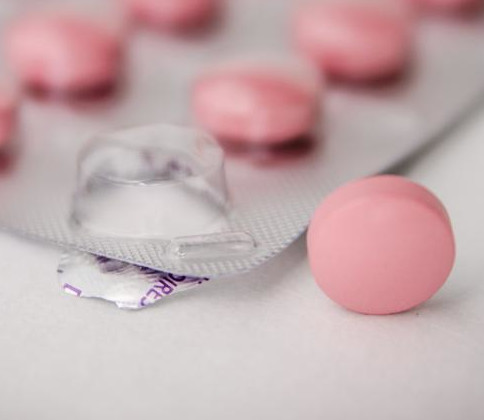}}{pill \\ triangle \\         circle}
  \jsubfig{\includegraphics[height=1.5cm]{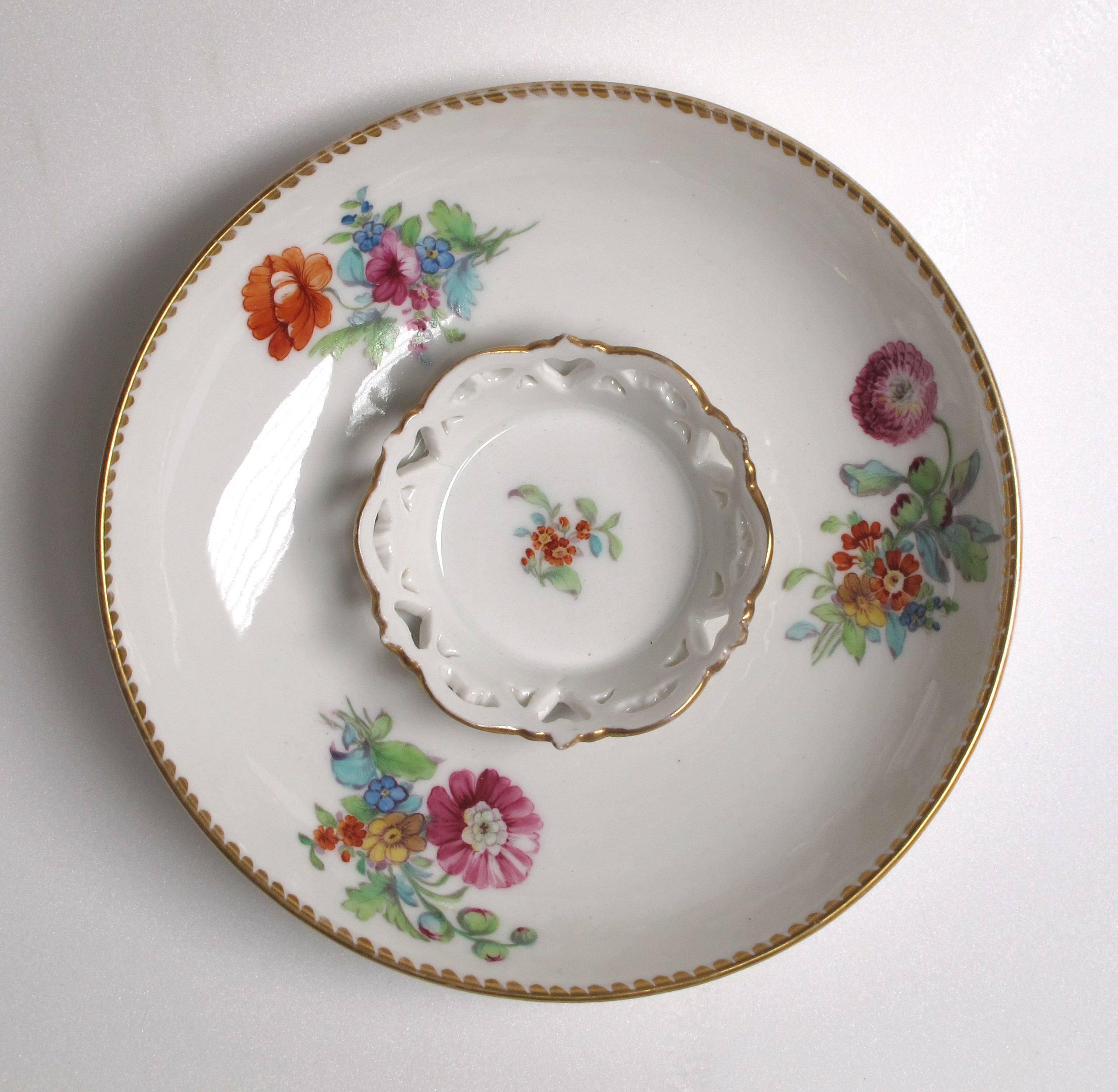}}{saucer \\ rectangle  \\        circle}
  \jsubfig{\includegraphics[height=1.5cm]{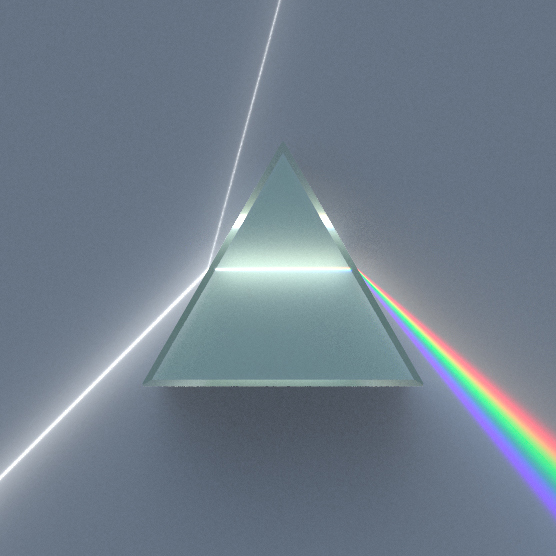}}{prism \\ triangle  \\        triangle}
  \jsubfig{\includegraphics[height=1.5cm]{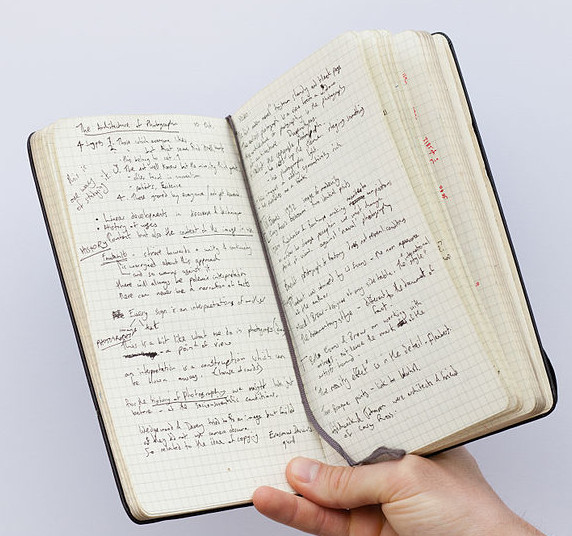}}{notebook \\ circle  \\        rectangle}
    \caption{\textbf{Examples of shape prediction results on ShapeIt benchmark}. Depicted above are examples of results for predicting shapes from the ShapeIt benchmark using MLM for BERT-base and Stroop probing for CLIP. Images are for illustration only, but during probing the model only receives the name of the item.}
    \label{fig:shapes}
\end{figure}

Figure \ref{fig:concreteness_examples} shows the results of concreteness prediction for CLIP and BERT. Nouns predicted as most concrete by CLIP, for example \emph{bench} and \emph{chalk}, that can be clearly visualized, while nouns predicted as least concrete (i.e., abstract) such as \emph{story} and \emph{name}, do not have a clear visual representation. In comparison, BERT's predictions are significantly noisier, with nouns such as \emph{seed} and \emph{jelly} predicted as abstract.

Figures \ref{fig:colors} and \ref{fig:shapes} shows color and shape association predictions of BERT-base and CLIP on samples from the relevant datasets. Without having access to the associated images, the CLIP text encoder usually predicts the correct matching between the given item and its correct shape or color, while BERT fails in most cases. Our results suggest that these associations are more consistently encoded by multimodally trained encoders. Furthermore, qualitative analysis of the misclassifications of CLIP, OpenCLIP and FLAVA on color association prediction reveals that these are mostly due to ambiguities in the dataset itself; see the supplementary materials for details.

Performance on non-visual NLU tasks, shown on the right side of Table \ref{tab:all_results}, demonstrates that our results are not an artifact of our probing methodology providing a global advantage to multimodally trained models, nor are these models uniformly better at language-related tasks. We also see that the non-visual tasks are highly solveable, with BERT-large and RoBERTa-large achieving high performance on all tasks despite the challenging zero-shot regime and limited information in the task inputs (ambiguity in cloze contexts for factual probing, lack of textual textual context for proficiency probing and randomly-chosen sentences for sentiment analysis). Despite this, the multimodally trained models show near-random performance.

We note a direct connection to the original Stroop effect in the field of human psychology. Follow-up works to the first Stroop effect demonstration have found it to apply to various types of stimuli, such as color-congruent and incongruent objects (e.g. a yellow banana vs. a purple banana)~\cite{naor2003color}. Our results, also including color congruence of objects, strengthen the motivation for using Stroop probing applied to tasks involving visual congruence or saliency. 

We also note a connection between our results and the \emph{reporting bias} effect, in which commonsense properties are less likely to be explicitly stated than incongruent properties (e.g. \emph{a (yellow) banana} vs. \emph{a blue banana}). Reporting bias in text has been studied in the context of color associations~\cite{paik2021world} and in more general contexts~\cite{shwartz2020neural,gordon2013reporting}. As the multimodally trained models under consideration were trained on paired image-caption data, the distribution of text in image captions differs somewhat from the text used for training models such as BERT. In the supplementary material, we provide an analysis of reporting bias in the LAION dataset~\cite{schuhmann2022laion}, the training data for the OpenCLIP model included in our tests. These results provide evidence that the improvement in performance seen from V\&L training cannot primarily be attributed to a lack of reporting bias in image caption texts, and emphasizes the significance of the visual modality in these models' language understanding.
\section{Conclusion}

We propose a suite of visual language understanding tasks along with non-visual natural language understanding tasks to probe the effect of V\&L pretraining on such reasoning capabilities of text encoder models. We introduce Stroop probing as a zero-shot knowledge proving method for evaluating the innate knowledge of text encoders. We also show that exposure to V\&L data in pretraining improves the performance of text encoder models on VLU tasks, even though they may underperform unimodally trained text encoders on non-visual NLU tasks.
Beyond text-only tasks, these results bear importance in the broader context of multimodal learning, in which the isolated contribution of text encoders has previously been underexplored. Our findings suggest that multimodal pretraining has a significant effect on the knowledge represented by the text encoder component of multimodal models, facilitating in establishing best practices for the design and training of text encoders used in such contexts.

\medskip \noindent \textbf{Acknowledgements.} 
We thank Noriyuki Kojima, Gabriel Stanovsky, and Adi Haviv for their helpful feedback.

{\small
\bibliographystyle{ieee_fullname}
\bibliography{egbib}
}

\clearpage
\appendix
{\LARGE\textbf{Appendix}}
\section{Additional Results and Comparisons}

\subsection{Non-dual V\&L encoder models}

\begin{table}[t]
  \centering
  \setlength{\tabcolsep}{3.2pt}
  \def\arraystretch{0.95}
  \begin{tabularx}{1.03\columnwidth}{lllcccccc}
    \toprule
    Task             &   & Metric                       & VisualBERT & LMXERT   &  BERT & CLIP        \\
    \midrule
    \textbf{VLU} \\
    Conc.            &   & Pearson           & 0.400 & 0.421    &  0.233        & 0.513          \\
                     &   & Spearman          & 0.412 & 0.370    &  0.238        & 0.495      \\
                     &   & Kendall           & 0.281 & 0.249   &  0.159        & 0.339      \\
    NCD              &   & Accuracy                     & 0.467 & 0.400  & 0.267    &  0.823\\
    CTD              &   & Accuracy                     & 0.314 & 0.431  & 0.353    &  0.800\\
    \midrule
    \textbf{NLU} \\
    Cites            &   & R@1                      & 0.003 & 0.007 & 0.199     & 0.019 \\
    NLI              &   & AUC                          & 0.704 & 0.688       & 0.754 & 0.696                     \\
    \bottomrule
  \end{tabularx}
  \vspace{3pt}
  \caption{Evaluating non-dual V\&L encoders (VisualBERT and LMXERT) on several VLU and NLU tasks with BERT and CLIP added for reference.} 
\label{tab:nondualencoderresults}
\end{table}

Although non-dual (fusion) encoder models are not directly comparable to purely textual encoders such as BERT or the text encoder component of CLIP which do not fuse modalities, we consider them here for completeness We evaluate the VisualBERT and LMXERT non-dual encoder models on several tasks from our task suite by only feeding them textual input. Results are shown in Table \ref{tab:nondualencoderresults}. As illustrated in the table, even though these models were trained using image features together with text tokens, the models outperform BERT on visual tasks, though the gap is not as significant as with dual encoder models.

\subsection{Additional tasks using linear probing}

We present two additional tasks for comparing V\&L and unimodal models using linear probing, one VLU and one NLU task. For both tasks, we use a linear classifier on the pooled embedding output of a model for categorical prediction. Specifically, we use a logistic regression model using the scikit-learn \texttt{linear\_model.LogisticRegression} implementation. For all tasks we use the default parameters except for \texttt{max\_iter} which was changed according to task requirements to allow convergence. In particular, we use parameters \texttt{penalty='l2', C=1.0, solver='lbfgs'}.

\subsubsection{Groundability classification}

\medskip \noindent \textbf{Task description.} In paired text-image data, there is normally an implied mapping between referential expressions in the text and objects or regions in the accompanying image. The task of learning these mappings is known as \emph{visual grounding} and is of general interest for visual semantic understanding \cite{Mao2015:refexp,Wang:16phraselocalize,cui2021s}. In captions accompanying images, some expressions refer directly to regions in images while others give non-visual context; we refer to the former as \emph{groundable} referents and the latter as \emph{non-groundable}. A similar paradigm was recently proposed by Kim et al.~\cite{kim2022flexible} that separately considers ``answerable'' and ``unanswerable'' phrases. 

We propose a \emph{groundability classification} task, consisting of classifying referents in text as groundable or non-groundable. This is a text-only task as it uses text alone and the visual context is only implied. Since this task requires visual imagination to complete, we consider it to be a VLU task.

\medskip \noindent \textbf{Experimental details.} In line with previous works that consider person-centric visual grounding \cite{cui2021s,qu2022weakly}, we construct a dataset of \emph{person-centric groundability} sentences where a fixed human participant is either implied to be groundable (i.e., on-camera) or non-groundable (i.e., off-camera). The associated task consists of binary classification applied to these texts according to whether the given participant would be visible in a description of an event. Due to the lack of existing labelled data for this task, we created a synthetic dataset of sentences with a common format: \textit{Alex \MASK{}ing Riley's \MASK{}}, where the first masked word is a randomly drawn verb, and the second masked word is a randomly drawn noun, and the task is to classify whether or not the second mentioned individual (i.e., Riley) is groundable. Groundability labels are estimated using zero-shot text classification with a pretrained natural language inference model. 
We created synthetic data for the groundability task by taking the prompt template ``Alex \rule{0.2in}{.5pt}ing Riley's \rule{0.2in}{.5pt}'', filling in various verb-noun pairs into the given slots, and filtering using a pretrained language model to select for natural-sounding samples. We then estimated ground-truth labels using zero-shot inference with a pretrained natural language inference (NLI) model.

To find verb-noun pairs, we listed all verbs and nouns in the Brown corpus of standard American English with part-of-speech labels~\cite{francis1964standard}. We converted all text to lowercase and then selected the 5,000 most common verb lemmas and 1,000 most common noun lemmas in this corpus. Using all possible verb-noun combinations among these, inserted into the prompt template shown above, yielded 5M candidate phrases. From the given 5M candidates we sample randomly 200K phrases. We then calculate the total negative log-likelihood (NLL) for each candidate relative to the pretrained language model GPT2-large~\cite{radford2019language} and kept only those samples in the 20th percentile of NLL (i.e. the most likely samples), corresponding to 40,000 descriptions.

After generating these texts, we estimated labels using a pretrained NLI model. We used BART-large~\cite{lewis2019bart} fine-tuned on the MNLI dataset~\cite{williams2017broad} (using the \texttt{facebook/bart-large-mnli} checkpoint from Hugging Face model hub\footnote{\href{https://huggingface.co/facebook/bart-large-mnli}{The model can be found here.}}). 
This model takes pairs of texts as inputs (the ``premise'' and ``hypothesis'' texts) and outputs three probabilities per pair: $p_c, p_n, p_e$, corresponding to probabilities of a contradictory, neutral, or entailment relation between the texts respectively. As observed by Yin et al.~\cite{yin2019benchmarking}, NLI can be used for zero-shot text classification by designing premise and hypothesis prompts for the task of interest. In our case, we use the following prompts:

\textbf{Premise:} ``This is a picture of \rule{0.5in}{.5pt}.''

\textbf{Hypothesis:} ``Riley can be seen in the picture.''

For each of our 40,000 texts, we insert the text in the slot given in the premise and calculate $p_e$ with the NLI model. If $p_e > 0.5$ we assign the sample label $1$ (groundable), otherwise we assign it label $0$ (non-groundable). 
Below are several example sentences from the synthetic dataset. 

\noindent Examples of Riley being groundable (sample label $1$): 
\begin{itemize}
    \item{Alex facing Riley's figure}
    \item{Alex viewing Riley's participation}
    \item{Alex seeing Riley's enjoyment}
\end{itemize}
Examples of Riley being non-groundable (sample label $0$): 
\begin{itemize}
    \item{Alex hiding Riley's file}
    \item{Alex announcing Riley's absence}
    \item{Alex stealing Riley's evidence}
\end{itemize}

For evaluation we created a test set, containing 200 sentences judged by human evaluators to be natural sounding, half labeled as groundable and the other half as non-groundable. To provide an example, sentences such as \emph{Alex cutting Riley's hair} or \emph{Alex blocking Riley's shot} were labeled as groundable, whereas sentences such as \emph{Alex painting Riley's house} or \emph{Alex counting Riley's vote} were labeled as non-groundable.

For this binary classification task, we apply linear probing to assess our models' understanding of groundability, and report ROC-AUC scores for each model. We also provide 95\% confidence intervals, calculated using bootstrap resampling with 200 bootstraps, in order to analyze the robustness of these results.

\begin{table}[t]
  \centering
  \setlength{\tabcolsep}{3.2pt}
  \def\arraystretch{0.95}
  \begin{tabularx}{0.52\columnwidth}{lll}
    \toprule
    Model             &   & AUC (95\% CI)          \\
    \midrule
    BERT          &   & 0.789 $\pm$ 0.0007 \\
    RoBERTa       &   & 0.799 $\pm$ 0.0005 \\
    ERNIE         &   & 0.766 $\pm$ 0.0006 \\
    CLIP              &   &\textbf{0.822 $\pm$ 0.0007}          \\
    \bottomrule
  \end{tabularx}
  \vspace{3pt}
  \caption{\textbf{Groundability Classification Evaluation}. We report ROC-AUC with 95\% bootstrap confidence intervals scores for a manually assembled test set comparing linear probing for text based encoders and V\&L CLIP model. As shown above, CLIP significantly outperforms the unimodally trained models.}
\label{tab:groundabilityresults}
\end{table}

\medskip \noindent \textbf{Results and discussion.} Results for the groundability classification are provided in Table \ref{tab:groundabilityresults}. As these results illustrate, CLIP significantly outperforms all unimodally trained text encoders on average. We observe that the score gaps are not as distinct as in the previous zero-shot tasks, as this task is a learnable task which requires training, allowing all models to learn this task to some extent. Nonetheless, CLIP's ability to surpass the unimodally trained encoders suggest that V\&L trained text encoders have a better ability to grasp if an object is grounded or not due to additional perceptual information that is encoded during the pretraining phase. Furthermore, note that in comparison to the other VLU tasks, here the subject in question (i.e., Riley) is not directly connected to visual information and the prediction is based only on context relating to the performed action and the associated object. The improved performance on this task illustrates that V\&L models can better encode higher-level perceptual reasoning. 

\subsubsection{Natural language inference}

\medskip \noindent \textbf{Task description.} Natural language inference (NLI) refers to inferring the logical relation between pairs of statements, as well as more generally referring to logical inference based on text~\cite{storks2019recent}. In particular, NLI commonly  considers the following logical relations between sentences A and B:

\begin{itemize}
    \item \textbf{Contradiction}: For example, A=\emph{It is rainy outside.} is contradicted by B=\emph{It is sunny outside.}, since they cannot be simultaneously true.
    \item \textbf{Neutral}: For example, A=\emph{It is rainy outside.} is neutral with regards to B=\emph{It is summer.}, since A neither contradicts nor entails B.
    \item \textbf{Entailment}: For example, A=\emph{It is cold and rainy outside.} entails B=\emph{It is cold outside.}, since if A is true then B must also be true.
\end{itemize}

Solving this task requires an understanding of the fine-grained semantics of language and logical reasoning. On the other hand, visual cues are not tightly related to this task and are even potentially misleading. For example, the sentences \emph{This cup contains grape juice.} and \emph{This cup contains wine.} are contradictory even though the scenes they describe are visually identical.  Therefore, we consider this to be a non-visual NLU task.

\medskip \noindent \textbf{Experimental details.} For this task we use the MNLI dataset introduced by Adina et al.~\cite{williams2017broad}. We remove sentence pairs with a neutral relation and treat this as a binary classification task to predict sentence pairs as contradictory or entailing. We perform 5-fold cross validation on a dataset of 261,775 pairs of sentences using 80\% of samples for training and 20\% for testing.

For each sentence pair, we concatenate the sentences' two pooled embeddings and apply linear probing. Note that some models such as BERT include a special \SEP token for encoding sentence pairs as a single unit, but we encode sentences separately and concatenate their embeddings in order to have a fair comparison between all models. We report the ROC-AUC score on the MNLI test set.

\begin{table}[t]
  \centering
  \setlength{\tabcolsep}{3.2pt}
  \def\arraystretch{0.95}
  \begin{tabularx}{0.48\columnwidth}{lccccccc}
    \toprule
    Model               &   & AUC $\pm$ std         \\
    \midrule
    BERT                &   & 0.754 $\pm$ 0.001     \\
    RoBERTa             &   & 0.777 $\pm$ 0.001     \\
    ERNIE               &   & \textbf{0.787} $\pm$ 0.001     \\
    CLIP                &   & 0.696 $\pm$ 0.001     \\
    \bottomrule
  \end{tabularx}
  \vspace{3pt}
  \caption{\textbf{NLI Evaluation}. We report ROC-AUC scores for the NLI task using linear probing, comparing text based encoders to the V\&L CLIP model. As depicted above, the V\&L trained text encoder is inferior to all other text based encoders for this non-visual language understanding task.}
\label{tab:nonvisualresults}
\end{table}

\medskip \noindent \textbf{Results and discussion.} Results for NLI are provided in Table \ref{tab:nonvisualresults}. As shown in the table, text-based models outperform CLIP by a large margin. Similar to our findings regarding linguistic acceptability classification, we see that V\&L trained models are less effective in tasks that do not incorporate perceptive information, suggesting that for non-visual tasks, V\&L pretraining is not necessarily beneficial. 

\subsection{Comparing usage of SP on text based models}

\begin{table*}[t]
  \centering
  \setlength{\tabcolsep}{3.5pt}
  \def\arraystretch{0.95}
  \begin{tabularx}{0.75\textwidth}{lcccccccccccccc}
     \toprule
     
    &
    \multicolumn{2}{c}{\textbf{Color}} &
    \textbf{Shape} & & & 
    \multicolumn{2}{c}{\textbf{Knowledge}} &
    \multicolumn{3}{c}{\textbf{Proficiency}} &
    \textbf{Sent.} 
    \\
    \cmidrule(lr){2-3}
    \cmidrule(lr){4-4}
    \cmidrule(lr){7-8}
    \cmidrule(lr){9-11}
    \cmidrule(lr){12-12}
    Metric &
    $acc_{\texttt{CTD}}$ & $acc_{\texttt{NCD}}$ & acc &&&
    R@1 & R@5 & $acc_\texttt{V}$ & $acc_\texttt{N}$ & $acc_\texttt{P}$ & acc
    \\
    \midrule
    BERT-MLM & \textbf{0.353} & \textbf{0.400} & \textbf{0.559} &&& \textbf{0.198} & \textbf{0.522} & \textbf{0.898} & \textbf{0.753} & \textbf{0.893} & \textbf{0.618} \\
    BERT-SP  & 0.137 & 0.067 & 0.412 &&& 0.000 & 0.005 & 0.048 & 0.038 & 0.013 & 0.596 \\
    \midrule
    RoBERTa-MLM & \textbf{0.431} & \textbf{0.333} & \textbf{0.431} &&& -- & -- & \textbf{0.877} & \textbf{0.718} & \textbf{0.881} & \textbf{0.666} \\
    RoBERTa-SP  & 0.176 & 0.200 & 0.422 &&& -- & -- & 0.016 & 0.019 & 0.063 & 0.616 \\
    \bottomrule
    
  \end{tabularx}
  \vspace{-5pt}
  \caption{\textbf{Comparing SP to MLM probing for text base models}. As the results show, using probing using MLM method for text based models outputs better results than using SP} 
\label{tab:comparingspmlm}
\end{table*}

In the main paper we presented results for text models using MLM probing, and for CLIP using Stroop probing (SP). To allow for a full comparison between both types of models, and to strengthen the choice of using MLM probing for text based models, we present additional results comparing SP and MLM probing for text based models. %
Table \ref{tab:comparingspmlm} presents results for comparing SP and MLM probing methods for BERT and RoBERTa. As illustrated, using SP with unimodally trained models results in lower performance than using MLM probing with these models. This result supports our choice of using MLM probing for text based models trained to perform MLM tasks as the preferred probing method. 

\subsection{Additional task results information}

\begin{table*}[t]
  \centering
  \setlength{\tabcolsep}{3.5pt}
  \def\arraystretch{0.95}
  \begin{tabularx}{0.94\textwidth}{lcccccccccccccc}
     \toprule
    &
    \multicolumn{3}{c}{\textbf{Concreteness}} &
    \multicolumn{2}{c}{\textbf{Color}} &
    \textbf{Shape} &&& 
    \textbf{Sent.} 
    \\
    \cmidrule(lr){2-4}
    \cmidrule(lr){5-6}
    \cmidrule(lr){7-7}
    \cmidrule(lr){10-10}
    Metric & $\left|\rho\right|$ & $\left|r_s\right|$ & $\left|\tau\right|$ &
    $acc_{\texttt{CTD}}$ & $acc_{\texttt{NCD}}$ & acc &&& acc
    \\
    \midrule
    \textbf{Unimodal} \\
    BERT-base & 0.27 $\pm$ 0.10 & 0.27 $\pm$ 0.09 & 0.18 $\pm$ 0.07 & 0.26 $\pm$ 0.13 & 0.25 $\pm$ 0.08 & 0.47 $\pm$ 0.08 &&& 0.56 $\pm$ 0.03 \\
    BERT-large & 0.18 $\pm$ 0.13 & 0.26 $\pm$ 0.10 & 0.17 $\pm$ 0.06 & 0.28 $\pm$ 0.14 & 0.27 $\pm$ 0.15 & 0.51 $\pm$ 0.06 &&& 0.56 $\pm$ 0.03 \\
    DistilBERT & -- & -- & -- & 0.23 $\pm$ 0.08 & 0.31 $\pm$ 0.04 & 0.45 $\pm$ 0.09 &&& 0.56 $\pm$ 0.04 \\
    RoBERTa-base & 0.30 $\pm$ 0.09 & 0.29 $\pm$ 0.10 & 0.19 $\pm$ 0.07 & 0.27 $\pm$ 0.10 & 0.27 $\pm$ 0.07 & 0.43 $\pm$ 0.00 &&& 0.61 $\pm$ 0.04 \\
    RoBERTa-large & 0.21 $\pm$ 0.10 & 0.23 $\pm$ 0.11 & 0.16 $\pm$ 0.07 & 0.30 $\pm$ 0.12 & 0.26 $\pm$ 0.08 & 0.43 $\pm$ 0.00 &&&  \textbf{0.63 $\pm$ 0.06} \\
    DistilRoBERTa & -- & -- & -- & 0.24 $\pm$ 0.12 & 0.25 $\pm$ 0.10 & 0.43 $\pm$ 0.01 &&& 0.57 $\pm$ 0.02 \\
    ERNIE & 0.23 $\pm$ 0.10 & 0.20 $\pm$ 0.12 & 0.13 $\pm$ 0.08 & 0.10 $\pm$ 0.04 & 0.13 $\pm$ 0.11 & 0.31 $\pm$ 0.08 &&&  0.53 $\pm$ 0.02 \\
    ERNIE-large &  0.23 $\pm$ 0.08 & 0.22 $\pm$ 0.07 & 0.14 $\pm$ 0.05 & 0.12 $\pm$ 0.06 & 0.07 $\pm$ 0.09 & 0.30 $\pm$ 0.05 &&& 0.57 $\pm$ 0.05 \\
    SBERT & 0.24 $\pm$ 0.09 & 0.25 $\pm$ 0.09 & 0.17 $\pm$ 0.06 & 0.13 $\pm$ 0.02 & 0.07 $\pm$ 0.01 & 0.43 $\pm$ 0.05 &&&  0.53 $\pm$ 0.02 \\
    \midrule
    \textbf{V\&L} \\
    CLIP & \textbf{0.47 $\pm$ 0.09} & 0.49 $\pm$ 0.09 & 0.34 $\pm$ 0.07 & 0.67 $\pm$ 0.15 & \textbf{0.70 $\pm$ 0.08} & 0.69 $\pm$ 0.08 &&& 0.52 $\pm$ 0.01 \\
    OpenCLIP & 0.45 $\pm$ 0.12 & 0.47 $\pm$ 0.12 & 0.32 $\pm$ 0.09 & \textbf{0.77 $\pm$ 0.12} & 0.66 $\pm$ 0.17 & \textbf{0.79 $\pm$ 0.08} &&& 0.53 $\pm$ 0.01 \\
    FLAVA & 0.46 $\pm$ 0.10 & \textbf{0.52 $\pm$ 0.10} & \textbf{0.36 $\pm$ 0.07} & 0.52 $\pm$ 0.30 & 0.47 $\pm$ 0.22 & 0.68 $\pm$ 0.10 &&& 0.50 $\pm$ 0.01 \\
    \bottomrule
    
  \end{tabularx}
  \vspace{-5pt}
  \caption{\textbf{Mean and STD Results.} Additional details of mean and standard deviations calculated across prompts, for all tasks which use multiple prompts.} 
\label{tab:allresultsmeanstd}
\end{table*}

We provide additional detailed results for our suite tasks including the mean and standard deviation of the results over all used prompts in Table \ref{tab:allresultsmeanstd}.

\subsection{Qualitative analysis for V\&L model misclassifications on color prediction}

\begin{table}[t]
  \centering
  \setlength{\tabcolsep}{3.2pt}
  \def\arraystretch{0.9}
  \begin{tabularx}{0.79\columnwidth}{llll}
    \toprule
    Word             &   & Ground Truth     & Predicted Color    \\
    \midrule
    apple           &   & green          & red               \\
    ash             &   & grey           & black             \\
    cauliflower     &   & white          & brown             \\
    cello           &   & brown          & black             \\
    chalk           &   & white          & grey              \\
    foam            &   & white          & grey              \\
    garlic          &   & white          & brown             \\
    lady finger     &   & green          & red               \\
    pear            &   & green          & yellow            \\
    sea             &   & blue           & grey              \\
    sky             &   & blue           & white             \\
    \bottomrule
  \end{tabularx}
  \vspace{3pt}
  \caption{\textbf{Qualitative results for CLIP misclassified objects from the CTD and NCD datasets.} As can be seen by analyzing the misclassified objects, most mistakes can be explained by ambiguity of the data.} 
\label{tab:clipmisclassification}
\end{table}

Our results for color association prediction show that V\&L models outperform unimodally trained text encoders in the given setting. Additional qualitative analysis of the results show that even the reported misclassifications of V\&L models such as CLIP may be explained by ambiguities in the dataset itself. For example, the noun ``ash'' has ground truth value ``grey'' in our dataset, while CLIP with SP predicts the color ``black'', which is arguably also correct. Table \ref{tab:clipmisclassification} presents all of the objects from both color datasets misclassified by CLIP, containing the ground truth and the predicted color. As seen there, most of these predictions may be interpreted as valid colors for the given objects.

\subsection{Analysis of reporting bias in LAION}

Prior works have noted that commonsense properties that can be inferred from text are less likely to be explicitly stated than incongruent properties, notably including color terms(e.g. \emph{a (yellow) banana} vs. \emph{a blue banana})~\cite{paik2021world,shwartz2020neural,gordon2013reporting}. In particular, text in image captioning datasets such as the web-scale LAION dataset\cite{schuhmann2022laion} (used to train OpenCLIP) might have a different incidence of reporting bias than the text used to train models such as BERT. To disentangle this from the effect of training on the visual modality, we provide an analysis of reporting bias in LAION for color associations.

We use the \texttt{laion-2B-en} subset of 2.33 billion English-language image-caption pairs in the LAION-5B dataset, and estimate reporting bias by searching for bigram pairs $(c, w)$ where $c$ is a basic color term\footnote{one of $\{$red, orange, yellow, green, blue, black, white, grey, brown$\}$} and $w$ is a unigram noun from our color association datasets (CTD and NCD). The empirical probability of color $c$ immediately preceding $w$ is $P(c|w) = n_{(c, w)} / n_w$, where $n$ indicates the number of instances of the given ngram, and the associated color estimates are $\hat{c}_w = \arg \max_c P(c | w)$. For these estimates, the corresponding accuracy scores on the unigrams in our datasets are $acc_{\texttt{CTD}} = 0.549$ and $acc_{\texttt{NCD}}=0.714$, significantly below the accuracies achieved by all of the multimodally trained models under consideration on these datasets for the color prediction task. We also provide qualitative examples in Table \ref{tab:reporting_bias} showing the effect of reporting bias for various common nouns from these datasets. These results provide evidence that multimodally trained models' strong performance on VLU tasks cannot be explained away as stemming from a lack of reporting bias in the texts used to train them.

\begin{table}[t]
  \centering
  \setlength{\tabcolsep}{3.2pt}
  \def\arraystretch{0.9}
  \begin{tabularx}{0.6\columnwidth}{llll}
    \toprule
    Word             &   & Ground Truth     & LAION    \\
    \midrule
    banana           &   & yellow          & green               \\
    cherry           &   & red          & black               \\
    orange           &   & orange          & red               \\
    soil           &   & brown          & red               \\
    swan           &   & white          & black               \\
    wood           &   & brown          & white               \\
    \bottomrule
  \end{tabularx}
  \vspace{3pt}
  \caption{\textbf{Reporting bias in the LAION dataset}, illustrated by unigram nouns from the CTD and NCD datasets, along with their ground truth colors and the most commonly preceding colors in LAION.} 
\label{tab:reporting_bias}
\end{table}

\section{Additional Details} 
\subsection{Models}
\begin{table*}[t]
  \centering
  \setlength{\tabcolsep}{3.2pt}
  \def\arraystretch{0.90}
  \begin{tabularx}{\textwidth}{llllclccc}
    \toprule
    Model family    & Size   & Pretraining  & Params    &  MLM head? & Checkpoint \\
    \midrule
    BERT~\cite{devlin2018bert}            & base  & text    & 110M   & Y       & \texttt{bert-base-uncased}    \\
    BERT~\cite{devlin2018bert}            & large & text    & 340M   &  Y      & \texttt{bert-large-uncased}   \\
    RoBERTa~\cite{liu2019roberta}         & base  & text    & 124M   &  Y      & \texttt{roberta-base}         \\
    RoBERTa~\cite{liu2019roberta}         & large & text    & 355M   &  Y      & \texttt{roberta-large}     \\
    ERNIEv2~\cite{sun2019ernie,sun2020ernie}           & base  & text  & 109M   &  \hspace{5px}Y$^*$       &  \texttt{ernie-2.0-base-en}   \\
    ERNIEv2~\cite{sun2019ernie,sun2020ernie}           & large & text  & 335M   &  \hspace{5px}Y$^*$       &  \texttt{ernie-2.0-large-en}   \\
    DistilBERT~\cite{sanh2019distilbert}               & base  & text  & 66M    &  Y       &  \texttt{distilbert-base-uncased} \\
    DistilRoBERTa~\cite{sanh2019distilbert}            & base  & text  & 82M    &  Y       &  \texttt{distilroberta-base} \\
    SBERT~\cite{reimers2019sentence}                   & --    & text  & 23M  &  N         &  \texttt{paraphrase-MiniLM-L6-v2} \\
    FLAVA~\cite{singh2022flava}                        & --    & text \& VLP  & 109M    &  Y        &  \texttt{facebook/flava-full}  \\
    CLIP~\cite{radford2021learning}                    & --    & VLP  & 63M    &  N        &  \texttt{openai/clip-vit-base-patch32}  \\
    OpenCLIP~\cite{ilharco_gabriel_2021_5143773}       & --    & VLP  & 352M    &  N        &  \texttt{laion/CLIP-ViT-H-14-laion2B-s32B-b79K}  \\
    \bottomrule
  \end{tabularx}
  \vspace{3pt}
  \caption{\textbf{Models table}. Note that the number of parameters listed for CLIP, OpenCLIP and FLAVA refers to their text encoder components alone. $^*$ Note: ERNIE was trained with an MLM head, but because the public checkpoints provided do not include this, we do not evaluate it with MLM probing.}
\label{tab:models}
\end{table*}

Table \ref{tab:models} presents the different models and Hugging Face checkpoints used for comparing results on the presented tasks.
\subsection{Prompts used per task}

We present further implementation details elaborating the list of prompts used per task.

\medskip \noindent \textbf{Concreteness Prediction}
As explained in the main paper, we use the following prompts to probe our models for the concreteness of words in context by using a cloze task paradigm with Stroop probing. For each word tested, we insert the masked prompt and the prompt with the tested word and calculate the cosine similarity between them.
\begin{itemize}
  \item \emph{Alice giving the $\SMASK$ to Bob}
  \item \emph{Bob giving the $\SMASK$  to Alice}
  \item \emph{I see the $\SMASK$}
  \item \emph{A photo of my $\SMASK$}
  \item \emph{A close-up photo of a $\SMASK$}
  \item \emph{A painting of the $\SMASK$}
  \item \emph{A photo of the $\SMASK$}
  \item \emph{A photo of a nice $\SMASK$}
  \item \emph{A drawing of the $\SMASK$}
\end{itemize}

\medskip \noindent \textbf{Color Association Prediction}
For the color association prediction, we use the following prompts. For each given object denoted as \W, we use all color options to probe for the correct color.
\begin{itemize}
  \item \emph{A picture of a $\SMASK$ \W}
  \item \emph{A photo of a $\SMASK$ \W}
  \item \emph{A photo of the $\SMASK$ \W}
  \item \emph{A $\SMASK$ \W}
  \item \emph{$\SMASK$ \W}
  \item \emph{The normal color of a \W is $\SMASK$}
  \item \emph{\W usually has a $\SMASK$ color}
  \item \emph{\W s have a $\SMASK$ color}
  \item \emph{What is the color of a \W? $\SMASK$}
  \item \emph{The natural color of a \W is $\SMASK$}
\end{itemize}

\medskip \noindent \textbf{Shape Association Prediction}
For the shape association prediction, we use the following prompts. For each given object denoted as \W, we use the given shape o to probe for the correct object shape.
\begin{itemize}
  \item \emph{A photo of a $\SMASK$ shaped \W}
  \item \emph{A photo of a $\SMASK$ \W}
  \item \emph{A photo of the $\SMASK$ \W}
  \item \emph{A $\SMASK$ \W}
  \item \emph{$\SMASK$ \W}
  \item \emph{An image of a $\SMASK$ \W}
  \item \emph{A \W usually has a $\SMASK$ shape}
  \item \emph{\W s commonly have a $\SMASK$ shape}
  \item \emph{The basic shape of a \W is $\SMASK$}
  \item \emph{What is the shape of a \W? $\SMASK$}
\end{itemize}

\medskip \noindent \textbf{Sentiment Analysis}
For sentiment analysis, we concatenate the following prompts to the given reviews and use the different options for sentiment prediction.
\begin{itemize}
  \item \emph{Is this review positive? $\SMASK$; Yes, No}
  \item \emph{Is this a good movie? $\SMASK$; Yes, No}
  \item \emph{I conclude the movie was $\SMASK$; good, bad}
  \item \emph{The film was $\SMASK$; good, bad}
  \item \emph{I had a $\SMASK$ time; good, bad}
  \item \emph{The following movie review expresses what sentiment? $\SMASK$; Positive, Negative}
  \item \emph{Sentiment expressed for the movie is $\SMASK$; Positive, Negative}
  \item \emph{The overall review of the film is $\SMASK$; good, bad}
  \item \emph{The movie was $\SMASK$; good, bad}
  \item \emph{This movie is $\SMASK$; good, bad}
\end{itemize}

\end{document}